\documentclass[conference]{IEEEtran}
\IEEEoverridecommandlockouts
\usepackage{cite}
\usepackage{amsmath,amssymb,amsfonts}
\usepackage{algorithmic}
\usepackage{graphicx}
\usepackage{textcomp}
\usepackage{xcolor}
\def\BibTeX{{\rm B\kern-.05em{\sc i\kern-.025em b}\kern-.08em
    T\kern-.1667em\lower.7ex\hbox{E}\kern-.125emX}}

\usepackage[acronym]{glossaries}    
\makeglossaries

\newacronym{em}{EM}{exact match}

\newacronym{dnn}{DNN}{deep neural network}
\newacronym{dl}{DL}{Deep Learning}
\newacronym{sota}{SOTA}{state-of-the-art}
\newacronym{nlp}{NLP}{natural language processing}    
\newacronym{sla}{SLA}{service level agreement}
\newacronym{slo}{SLO}{service level objective}
\newacronym{qps}{QPS}{query per second} 
\newacronym{lm}{LM}{language model} 
\newacronym{cv}{CV}{computer vision} 
\newacronym{dag}{DAG}{directed acyclic graph} 
\newacronym{mcs}{MCS}{Monte Carlo search} 
\newacronym{nll}{NLL}{negative log likelihood} 
\newacronym{flop}{FLOP}{floating-point operation} 
\newacronym{flops}{FLOPS}{floating-point operations per second} 
\newacronym{moe}{MoE}{mixture-of-experts} 

\newacronym{sys}{MIE}{Mixed Inference Experts} 
\usepackage{xurl}
\usepackage{xspace}
\usepackage{multirow}
\usepackage{multicol}
\usepackage{subcaption}
\usepackage{bbm}
\usepackage{comment}
\usepackage[ruled,vlined, linesnumbered]{algorithm2e}

\newcommand{\review}[1]{\textcolor{black}{#1}}

\newcommand{\sys}{\textsc{HybridServe}\xspace}

\usepackage{tikz}
\newcommand*\myc[1]{%
\scalebox{0.78}{\begin{tikzpicture}[baseline=-3pt]
  \node[draw,circle,inner sep=0.5pt, fill=black] {\textcolor{white}{\textsf{\textbf{#1}}}};
\end{tikzpicture}}}

\usepackage{xcolor}

\newcommand{\tinyskip}{\vspace{3pt}}
\newcommand{\mypar}[1]{\tinyskip\noindent\textbf{#1.}\xspace}

\newcommand{\para}[1]{\noindent {\bf #1}}

\newcommand{\eg}{\text{e.g.,}\ }

\newcommand{\ie}{\text{i.e.,}\ }

\newcommand{\score}{C}
\newcommand{\energy}{e}

\newcommand{\utilization}{u}
\newcommand{\transmission}{\mathcal{T}}
\newcommand{\latency}{l}

\newcommand{\memory}{\omega}
\newcommand{\model}{m}

\newcommand{\outputs}{P}
\newcommand{\confidence}{f}
\newcommand{\scale}{\theta}

\newcommand{\eval}{{D_v}}

\newcommand{\labels}{Y}

\newcommand{\llikelihood}{\mathcal{L}}

\newcommand{\norm}[1]{\lr\lVert\rVert{#1}}
\newcommand{\softmax}[1]{\sigma\lr(){#1}}

\newcommand{\Min}[1][]{\min\limits_{#1}}
\newcommand{\Max}[1][]{\max\limits_{#1}}

\newcommand{\Sum}[3]{\sum\limits_{#1}^{#2}{#3}}

\newcommand{\lr}[3]{\left #1 {#3} \right #2}

\begin{document}

\title{\sys: Efficient Serving of Large AI Models with Confidence-Based Cascade Routing}

\author{\IEEEauthorblockN{Leyang Xue, Yao Fu, Luo Mai, Mahesh K. Marina}
\IEEEauthorblockA{\textit{The University of Edinburgh}}
}


\maketitle

\begin{abstract}
Giant Deep Neural Networks (DNNs), have become indispensable for accurate and robust support of large-scale cloud based AI services. However, serving giant DNNs is prohibitively expensive from an energy consumption viewpoint easily exceeding that of training, due to the enormous scale of GPU clusters needed to hold giant DNN model partitions and replicas. 
Existing approaches can either optimize energy efficiency or inference accuracy but not both. To overcome this status quo, we propose \sys, a novel hybrid DNN model serving system that leverages multiple sized versions (small to giant) of the model to be served in tandem. Through a confidence based hybrid model serving dataflow, \sys prefers to serve inference requests with energy-efficient smaller models so long as accuracy is not compromised, thereby reducing the number of replicas needed for giant DNNs. \sys also features a dataflow planner for efficient partitioning and replication of candidate models to maximize serving system throughput. Experimental results using a prototype implementation of \sys show that it reduces energy footprint by up to 19.8x compared to the state-of-the-art DNN model serving systems while matching the accuracy of serving solely with giant DNNs.  
\end{abstract}


\let\svthefootnote\thefootnote
\newcommand\freefootnote[1]{%
  \let\thefootnote\relax%
  \footnotetext{#1}%
  \let\thefootnote\svthefootnote%
}

\freefootnote{\hrule\vskip 3pt Short version of this paper to be presented at ICDCS 2025..}

\section{Introduction}

The increasing adoption of giant Deep Neural Networks (DNNs), especially transformer models with attention mechanisms, such as ViT~\cite{vit} and Llama~\cite{llama}, has made efficient serving of large models a critical focus for AI service providers. Modern DNN-serving systems, including DeepSpeed~\cite{deepspeed} and Triton~\cite{tritonserver}, typically rely on dedicated GPU server clusters operating 24/7 to handle model serving tasks. These systems divide large models into partitions, which are executed using automated model parallelism~\cite{alpaserve}. To optimize performance, request routers aggregate incoming user queries into batches and initiate inference on a per-batch basis.

However, the reliance on large-scale DNNs for production AI services incurs significant energy costs. Due to the limited throughput of individual model replicas, serving systems often deploy numerous replicas to handle high query volumes efficiently. This approach necessitates the allocation of large GPU server clusters running continuously, significantly increasing the operational energy footprint.
For example, serving GPT-3 at 1000 queries per second on NVIDIA T4 GPUs is estimated to consume $4.4\times10^{10}$J daily, or $1.6\times10^{13}$J annually—an order of magnitude higher energy expenditure over the model's lifetime compared to training.


To improve energy efficiency, DNN serving systems can compress~\cite{gptq} or distill~\cite{cot-distill} giant DNNs. Although model compression and distillation techniques
can significantly reduce the number of GPUs required for serving, they do so at the expense of reduced inference accuracy~\cite{DBLP:conf/iclr/ZhuG18}, hindering their widespread adoption. The alternative approach is to keep the model as it is but optimize the GPU runtime efficiency (\eg DeepSpeed~\cite{deepspeed}), only marginally reducing the number of GPUs required.


\review{
In this paper, we aim to achieve general purpose energy-efficient serving of giant DNNs without extra model fine-tuning.
Our key idea is to develop routing mechanisms that leverage the multiple released sizes of the same model from small to giant, differing in their energy footprint and inference accuracy.
For example, Google T5~\cite{DBLP:journals/jmlr/RaffelSRLNMZLL20} is released with 4 versions ranging from 300MB to 10GB in size.
These multiple versions strike different trade-offs between energy efficiency and inference accuracy~\cite{DBLP:journals/pieee/DengLHSX20} -- larger models yield higher inference accuracy but are more energy hungry; smaller models, on the other hand, exhibit relatively lower inference accuracy but are energy efficient.
In our proposed approach, the routers allow smaller models to answer most requests and only let the ones they have low confidence inference to be propagated further to be handled by larger models, thereby reducing the overall energy footprint of serving while yielding equivalent accuracy.
To realize the above outlined design, two main challenges need to be addressed: (i)~How to construct a hybrid model serving graph and compute the confidence for each model? (ii)~How to parallelize such a dataflow over distributed GPUs so that such a dataflow can meet the serving performance requirements (\eg latency and throughput)?
}

\review{
Our design idea has led to \sys, a serving system that can significantly reduce its deployment cost via offloading request to smaller models and effectively mitigate the need for large number of GPUs with small and giant models being packed on the same set of GPUs. \sys achieves this via the following key contributions:
}

\mypar{(1) Confidence based hybrid model serving dataflow}
\review{
We propose the confidence based hybrid model serving dataflow. The objective of this dataflow is to optimize the energy efficiency of serving with a group of candidate models (\eg small, medium and large models) while maintaining inference accuracy similar to that with the largest (giant) model.
}

\review{
The feasibility of such method is based on the following key observations:
(i) the smaller models' capability is not strictly the subset of the larger models, resulting in all model capability combined is better than one single large model.
This is primarily due to each sized model acting as an expert in a certain sub-domain during the inference.
This enables energy-accuracy trade-off space beyond the capabilty of the largest model;
(ii) a well-tuned small model can handle a majority of tasks, resulting in majority of inference requests being offloaded to smaller models.
This provides lower latency and higher energy efficiency for each request on average.
}

\review{
These observations enable a post-hoc collaboration between models while keeping the model architecture and parameter unchanged from user provided checkpoints, using calibration functions as plugins.
Leveraging the ability of DNNs to generalize, \sys learns a \emph{DNN confidence score function} and adapts it based on the inference task type (\eg classification, generation, or question answering). 
\sys then generates a \emph{threshold performance graph}, which predicts inference accuracy and energy efficiency at varying confidence thresholds. Users can select thresholds based on their serving priorities (\eg accuracy-oriented or energy-efficient). Additionally, \sys introduces \emph{skip connections} in the dataflow, enabling requests to bypass less confident models and reach confident ones directly, thereby speeding up processing in the hybrid model serving pipeline.
}

\mypar{(2) Hybrid serving dataflow planner}
\review{We propose dataflow planner that co-locates memory-capacity-bounded models (larger DNNs with low request rate) with compute-bounded models (smaller DNNs with high request rate) for higher GPU resource utilization.}
\sys parallelizes a hybrid model serving dataflow on distributed GPUs through a dataflow planner.
This planner partitions giant DNNs based on the memory capacity of a GPU and
multiplexes DNNs (and their partitions) onto GPUs with an aim of optimizing the dataflow's throughput in processing model serving requests.
To prevent DNNs from contending for resources, we propose fine-grained GPU occupancy metrics (\ie kernel utilization). Based on the metrics, the planner can ensure co-located DNNs can process concurrent requests in real-time. Moreover, \sys adaptively creates replicas for potential bottleneck DNNs in the dataflow so that the aggregated throughput is maximized.
Finally, the planner solves the planning problem in polynomial time, making \sys easy to be deployed in large-scale DNN serving clusters.


Experimental results on an 8-GPU server show that \sys can preserve the same level of accuracy as giant model with state-of-the-art giant neural networks -- ViT~\cite{vit}, T5~\cite{DBLP:journals/jmlr/RaffelSRLNMZLL20} and GPT-3~\cite{gpt3}. Compared to model compression techniques (\ie model distillation and quantization), \sys can achieve up to 2x better inference accuracy. 
We further evaluate \sys on a commercial cloud computing platform using a cluster of 12 servers (each with an NVIDIA T4 GPU). Compared to state-of-the-art high-performance distributed ML model serving systems -- DeepSpeed~\cite{deepspeed}, Triton~\cite{tritonserver} and Ray Serve~\cite{ray}, \sys can reduce energy by up to 19.8x, measured using the novel \emph{joule per request} metric (or equivalently, provide 8x higher throughput) when serving a synthetic GPT-3 model with around 20 billion parameters.

\section{Background and Motivation}

\subsection{DNN serving systems and energy costs}
\label{systems-energy-costs}

\begin{figure}[t!]
\centering
    \includegraphics[width=.9\linewidth]{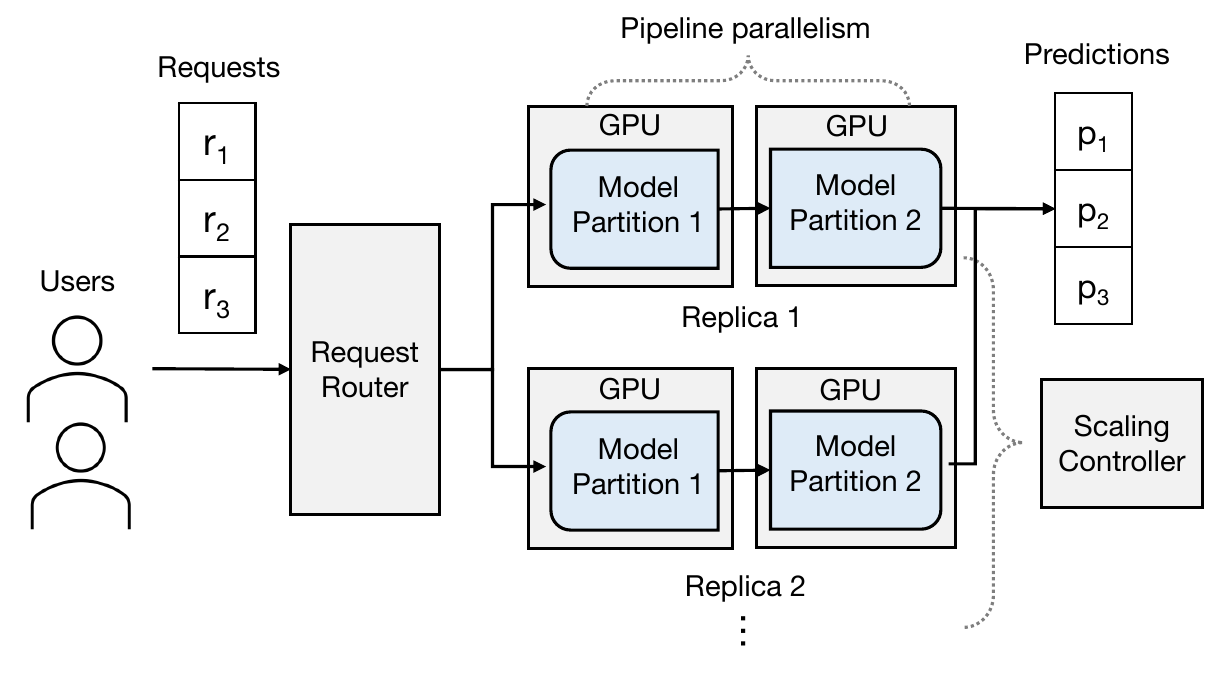}
    \vspace{-0.2in}
    \caption{Schematic of a typical giant DNN serving system.}
    \label{fig:serving-cluster}
\end{figure}

Several systems have been built to serve DNNs. These model serving systems typically follow an architecture shown in Figure~\ref{fig:serving-cluster}. In every time window, \emph{users}
produce a set of \emph{requests} (inference queries) that are sent to a \emph{request router}.
The router dispatches those requests across multiple replicas
of the DNN model. There is a \emph{scaling controller} which controls the number of DNN model replicas so that all requests can
be processed within their latency requirement (usually hundreds of milliseconds~\cite{orca,vllm}). Requests are processed in parallel by the different replicas and their combined set of \emph{predictions} are eventually returned to the users. When DNN models are large
and cannot be fit into a single GPU, the serving system
needs to partition these DNNs across multiple GPUs and processes each request 
in a \emph{model parallelism} manner~\cite{alpaserve}.

Serving giant DNNs with billions or trillions of parameters for production-grade AI services is extremely energy hungry.
Serving a giant DNN such as GPT-3 following the current system architecture shown in Figure~\ref{fig:serving-cluster} needs numerous GPUs (\eg NVIDIA T4) to be reserved 24/7, which causes high energy consumption.
To appreciate this, consider production-grade AI services (\eg image classification, question answering)
which have to serve hundreds or even thousands of requests per second (RPS)~\cite{clipper,orca}.
Since each giant DNN model can process tens of requests per second, DNN serving systems
have to create hundreds of model replicas, in turn requiring thousands of GPUs reserved 24/7.

We estimates the energy cost of serving a GPT-3 model with 100 billion parameters.
Training such a model consumes $2\time10^{14}$ Joules~\cite{gpt3-moon}, which is already enormous. In contrast,
serving such a model for a small-scale AI service with 100s  of RPS consumes $10^9$ joules per day, and after 900 days, the serving related energy consumption cost would exceed that of training.
Considering a medium-scale AI service with 1000s of RPS, the serving cost will jump to
$10^{10}$ joules per day and will surpass the training cost in only 100 days.
For a large-scale AI service~\cite{clipper} with an order of magnitude higher (tens of 1000s) RPS, it would take just a day for serving related energy cost to match the training cost. As such, it would be prohibitively expensive to incur the energy cost equivalent to model training for each additional day of serving. 


\subsection{Issues with prior energy-saving methods}

Existing approaches to improve energy efficiency of DNN serving systems fall into two categories, as discussed below.
\mypar{(1) Model compression} Several techniques exist to compress the size of (giant) DNNs:
(i) \emph{Knowledge distillation techniques} (\eg DistiLLM~\cite{distillm}, FCD~\cite{fcd}, AKD~\cite{DBLP:conf/icml/TouvronCDMSJ21}) train smaller DNN models by teaching them the behavior of larger ones; 
(ii) \emph{Quantization techniques} (\eg BitsAndBytes~\cite{bits-and-bytes} and GPTQ~\cite{gptq}), trade-off the precision of model weights for faster DNN inference;
(iii) \emph{Pruning techniques} (\eg LayerMerge~\cite{layer-merge}, Movement pruning~\cite{DBLP:conf/nips/Sanh0R20}) prune model parameters that would have minimal effect on inference accuracy. 
Compression techniques often treat model as a white box with its parameter and architecture modifiable. This often comes with extra finetuning rather than off-the-shelf deployment.
In addition, model generalizability can be compromised~\cite{DBLP:conf/iclr/ZhuG18}.

\mypar{(2) High-performance DNN serving systems}
Ray Serve~\cite{ray} and Clipper~\cite{clipper} allow multiple DNNs to effectively share GPUs, whereas Clockwork~\cite{clockwork} coordinates multiple DNNs based on monitored latency performance. 
These approaches work with giant DNN models as it is and so preserve accuracy but only achieve limited improvement in energy efficiency because of the large number of DNN replicas to maintain.
Serving systems like ServerlessLLM~\cite{serverlessllm} and SGLang~\cite{sglang} provide limited energy savings due to the continued need for large numbers of replicas for a single model.
Other system efficiency optimizations like GPU multiplexing~\cite{orion}, memory offloading~\cite{moe-infinity} and kernels~\cite{moe-gen,deepspeed} are complementary to \sys, as these choices consider pre-determined factors that are accounted for in profiling.

\mypar{(3) Model cascading}
Model cascading has been extensively studied in the context of edge cloud collaboration.
Systems like DCCL~\cite{DBLP:conf/kdd/YaoWJHZY21} rely on online training to adapt to user requests in the recommendation context.
DDNN~\cite{DBLP:conf/icdcs/Teerapittayanon17} and PerDNN~\cite{DBLP:conf/icdcs/JeongLSYM20} dynamically partition a giant DNN to balance request latency and dependency on cloud GPUs.
These still suffer from large resource usage.
Pregate methods like NoScope~\cite{noscope} and TAHOMA~\cite{DBLP:conf/icde/AndersonCRW19} train a separate router before all models to approximate the confidence score.
While effective for vision tasks, they struggle with language inputs as the processing is heavily context-based, while not raw inputs. 
RouteLLM~\cite{routellm} resolves the challenges in language models by training models with each other for cascading and HybridLLM~\cite{hybridllm} trains a small model from scratch to route user requests, while they do not scale with long cascading pipelines and off-the-shelf model serving.

\section{\sys Overview}

\subsection{Key Observations}

\review{
We have the following key observations that motivate our design to improve inference via offloading requests to smaller models.
We conduct a simple experiment with 3 sizes of the BERT model and considering the tasks in the GLUE dataset as Table~\ref{tab:collaborate-inference}.
For each task, we record for each test sample, whether each model produced a correct answer.
The joint accuracy represents an ideal post-hoc case where the accuracy with each sized model is combined together to yield the overall accuracy.
This also indicates the upper bound of request offloading and overall accuracy.
Such phenomenon is also widely observed among other language model and datasets such as GPT and Llama with MMLU~\cite{mmlu}, and can be extended to vision models like ViT.
Due to space limits, we omit the results.
}

\begin{table}[t]
    \centering
    \caption{Ideal accuracy comparison on GLUE tasks between a large BERT model and a set of different sized BERT models.}
    \begin{tabular}{|c|ccccc|}
    \hline
       Model  & CoLA & QQP & SST-2 & QNLI & RTE \\
    \hline
        BERT-large & 61\% & 72\% & 95\% & 93\% & 70\% \\
    \hline
       Joint Accuracy & 88\% & 98\% & 97\% & 94\% & 82\% \\
    \hline
    \hline
        \multicolumn{6}{|c|}{Joint Accuracy (contribution of each sized model)} \\
    \hline
        Distil-BERT (Small) & 82\% & 53\% & 55\% & 87\% & 64\% \\
        BERT-base (Medium)& 5\% & 44\% & 40\% & 6\% & 15\%     \\
        BERT-large (Large)& 1\% & 1\% & 2\% & 1\% & 3\%     \\
    \hline
    \end{tabular}
    \label{tab:collaborate-inference}
\end{table}

\para{(1) Grouped models are better than one.}
\review{
Our first finding is that joint accuracy is much higher than single large model with up to 27\% difference for the CoLA task in Table~\ref{tab:collaborate-inference}.
Given this is a case for three models combined, the improvement in accuracy comes from the smaller two models.
This indicates that the capability of giant DNNs is not strictly a superset of smaller equivalents.
Models being trained as sub-domain experts is also observed in the context of ensemble learning and mixture of experts~\cite{DBLP:conf/nips/DingC023,read-me}.
}
\review{
The above idea indicates models without dedicated training can also be sub-domain experts.
A strategy of discovering the model expertise in the post-hoc manner can improve overall inference accuracy.
}

\para{(2) High capability of small models.}
\review{
Next finding indicates that capability of smaller models is largely overlapped with the giant model.
The accuracy breakdown in the joint case in Table~\ref{tab:collaborate-inference} is constructed as follows: we first assess the responses to inference requests using the smallest model, and only the requests yielding incorrect answers are passed on to the next larger model and so on.
We observe that the smallest model can correctly handle over 80\% of the requests, ideally.
In addition, the giant model only needs to process 1-3\% of the overall requests when we first use smaller models.
}
\review{
Leveraging such property can significantly enhance the per-request energy consumption.
The smallest model is often more than 4$\times$ energy efficient then the giant model.
By offloading requests to such models, we can have at least 2$\times$ energy efficiency.
}


\subsection{System Design}

\begin{figure}[t]
\centering
    \includegraphics[width=\linewidth]{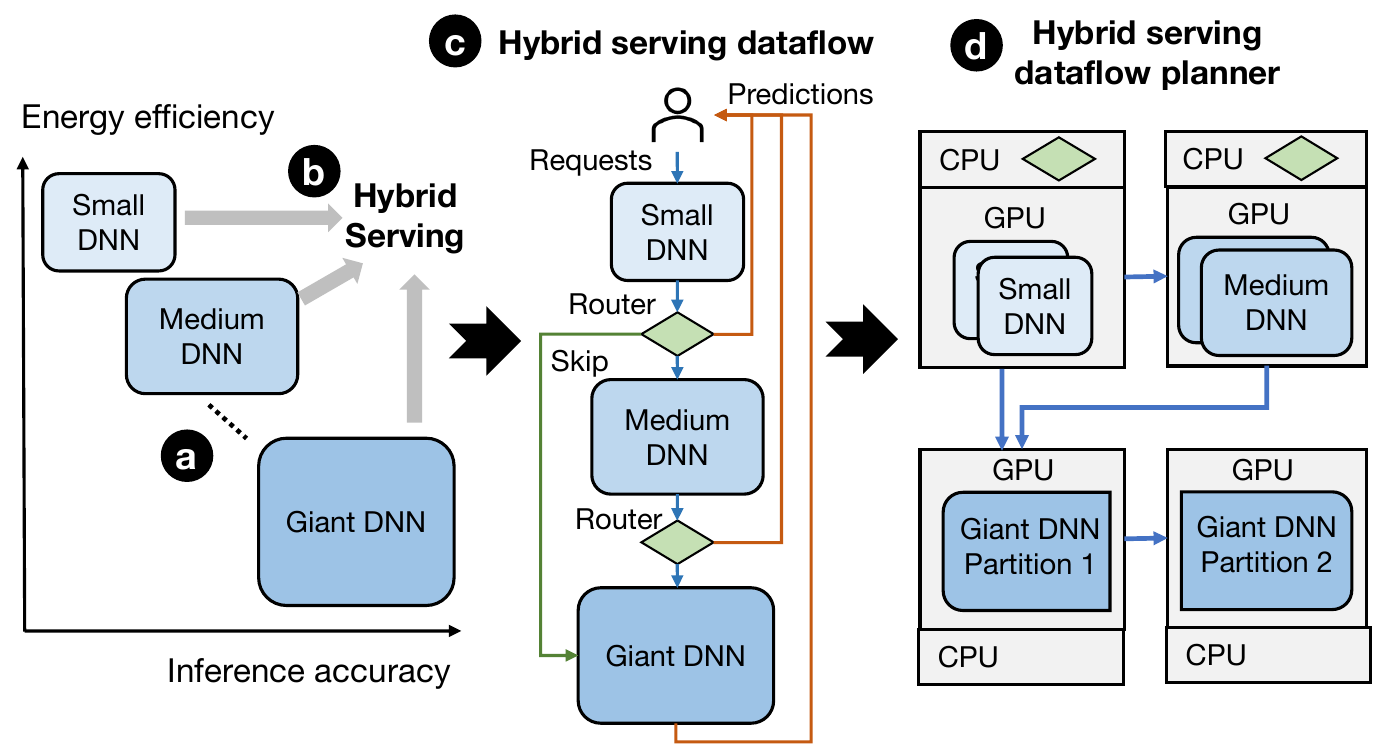}
    \caption{\sys system design overview.}
    \label{fig:DNN-landscape}
\end{figure}

Given the above observation, our aim to enable a hybrid DNN serving system that harnesses different sized DNNs,
which are often released by DNN providers (\eg HuggingFace~\cite{huggingface}) ranging from small, medium to giant to balance between energy efficiency
and inference accuracy. (as shown by \myc{a} in Figure~\ref{fig:DNN-landscape}). 
Such a hybrid serving system (\myc{b}) leverages small, medium, and giant DNNs to achieve both high energy efficiency and inference accuracy. This approach is analogous to hybrid cars, which optimize fuel efficiency by using an electric motor at low speeds and a combustion engine at higher speeds. Since low-speed operation dominates (e.g., urban commutes), the combustion engine is rarely needed. Similarly, our system primarily relies on smaller models, resorting to larger models only when necessary to preserve accuracy.

We design the \sys system to realize the above idea. It brings together two main aspects: (i) \emph{hybrid model serving dataflow} (\myc{c}) and (ii) \emph{hybrid serving dataflow planner} (\myc{d}). On the first aspect, the dataflow consists of nodes that correspond to different sized versions of the DNN model to be served; these nodes are interconnected using request routers. The routers in \sys are associated
with {\em confidence thresholds} that allow smaller models to serve inference queries when they are confident and only letting the remaining queries to reach larger models, thereby reducing the number of replicas required for giant DNNs. The routers are additionally associated with skip connections, which offer the option of requests directly
reaching the most confident models without incurring extra request routing overhead and latency penalty. Concerning the second aspect, the dataflow planner partitions giant DNNs based on the memory capacity of GPUs; it further replicates DNNs that can become bottlenecks on the dataflow (\eg the small and medium sized DNNs shown in Figure~\ref{fig:DNN-landscape}). Moreover, the planner places DNNs onto GPUs and routers onto CPUs with the aim of optimizing the aggregate throughput of processing model serving requests.

\section{Hybrid Model Serving Dataflow}
\label{sec:confidence}


\subsection{Learning DNN confidence score functions}
\label{sec:confidence-framework}



\review{
Confidence scores estimate the likelihood of a model's predictions being correct. However, generating reliable confidence scores is challenging because: (1) Ground truth is unavailable during inference, so confidence must be self-estimated; (2) Fine-tuning models for confidence estimation is costly and impractical, as serving models can be black boxes; (3) Models have heterogeneous prediction formats. To address this, we aim to create a mapping that derives reliable confidence scores from model predictions without altering model parameters or requiring ground-truth labels.
}

\begin{figure}[t]
    \centering 
    \includegraphics[width=\linewidth]{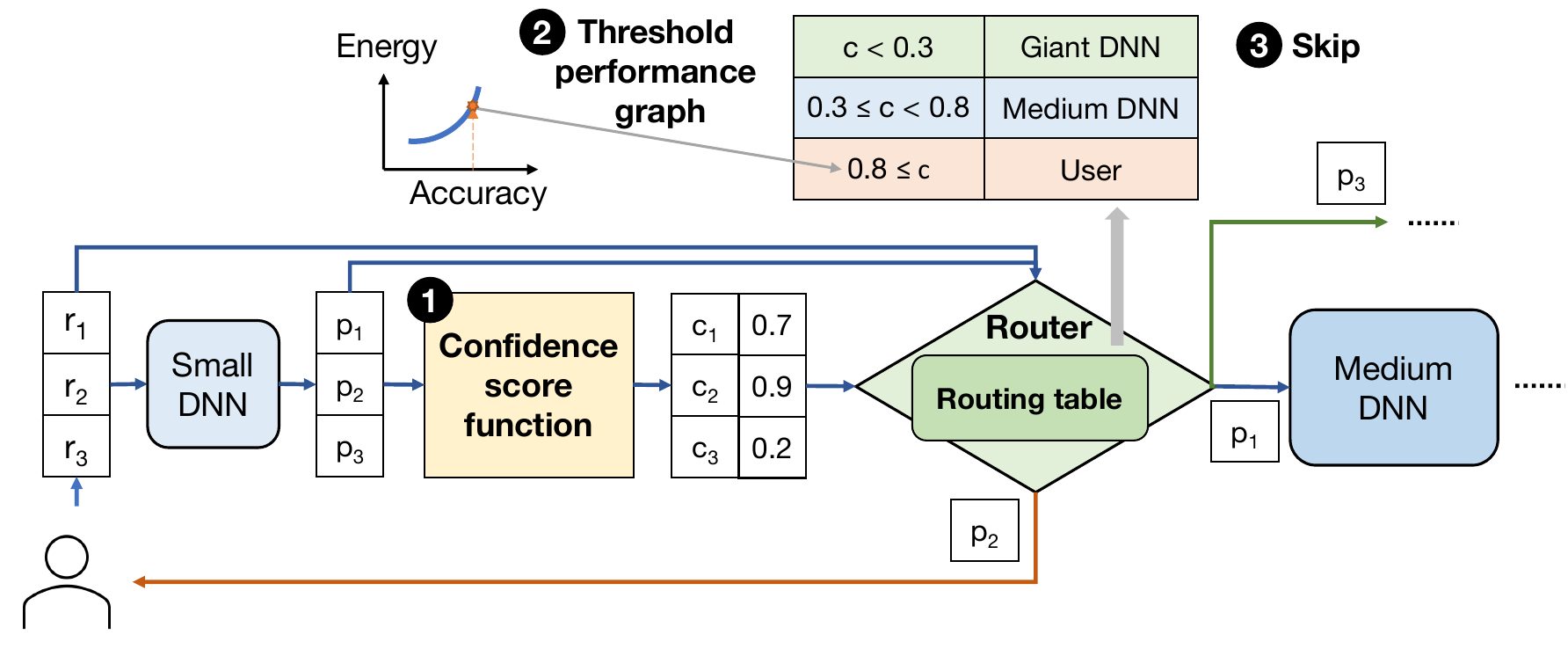}
    \vspace{-0.2in}
    \caption{Overview of hybrid model serving dataflow.}
    \label{fig:confidence-overview}
\end{figure}

Our key idea is to fit an additional layer as the \emph{confidence score function} (as shown by \myc{1} in Figure~\ref{fig:confidence-overview}) that learns the mapping between the model predictions $P$ and labels $Y$ based on a small validation dataset and to apply it on test/inference data. 
\review{We aim to apply a cost efficient yet robust calibration for each model \eg Temperature Scaling (TS)~\cite{tempscaling}.
We aim for the following property of the method:
(i) accuracy preserving: TS is essentially linear scaling of the model's original output distribution, which does not alter any model output, providing guarantee that the users' models are served as it is.
This is in contrast to post-hoc learning based approaches~\cite{DBLP:conf/nips/Lakshminarayanan17} that do not have such property.
(ii) system flexibility: compared to pregate approaches where gates need to be retrained if models are added or dropped from the dataflow~\cite{noscope},
substitutions for TS can be applied while empirically yielding the same result.
}
By doing so, as shown in Figure~\ref{fig:confidence-overview}, confidence scores (\ie $\{c_1, c_2, c_3\}$) can be estimated only based on the model predictions (\ie $\{p_1, p_2, p_3\}$). 


We design a unified framework to transform various model predictions into distinguishable confidence scores.
Given the model predictions $\outputs$, a confidence score function $f$ with parameters $\theta$ maps $\outputs$ to \emph{confidence scores} $C$, \ie $f_\theta: \outputs \mapsto \score$.
The confidence learning objective is to learn parameters $\theta$ that minimize the \acrfull*{nll} between confidence scores and labels $\labels$ on the validation dataset, \ie
\begin{align}
     \min_\theta\llikelihood\left(f_\theta(\outputs) | \labels\right), \label{eq:confidence}
\end{align}
where $\llikelihood$ is naturally cross entropy loss due to the discrete property of $\outputs$.

A key observation enables the unified framework -- the prediction formats of classification, generation, and question answering model can be seen as variants of classification outputs. Classification models output a floating-point vector, namely \emph{logits}, as a prediction for each input, whose length equals the number of classes. Each position in the logits indicates the probability that the input belongs to the corresponding class. 

In the following, we describe how to adapt this confidence score function for different prediction tasks:

\para{(1) Classification.} Text classification and image classification tasks have the same format of output, where the shape of the output vector is the number of classes. The function accepts both raw logits outputs from the model or outputs after the Softmax function.
Formally, $
f_\scale(\outputs) = \max\lr(){\theta(\outputs)^2}
$ if $\outputs$ is logits, otherwise $f_\scale(\outputs) = \theta\lr(){\Max(\outputs^2)}$.


    
\para{(2) Generation.} Generation models predict the next word from a vocabulary. The outputs of the generation task can be seen as classification across all words in the vocabulary. 
The vocabulary size of the generation model is usually in the order of $10^4$, which makes the strongest prediction less significant.
We assign $\outputs = \text{TopK}\lr(){\outputs}$ to enhance the prediction, while the rest is the same as classification task, \ie plugin in $\outputs$ above. 
During each iteration, the confidence of a token $\omega$ can be collected, among which the minimal confidence is the confidence of complete output.
Formally, $\confidence_\scale(\outputs) = \text{TopK}\lr(){\softmax{\scale(\outputs)}^2}$.


\para{(3) Question Answering.} Question answering tasks extract the exact text in the given context for a specific question. This can be seen as a binary classification task for each token, whether it is part of the answer or not. 
As such, minimal confidence is the confidence of the answer.
When either the confidence of the start or end token is below a threshold, the context and the question are passed on to the larger model for inference. 
Formally,$\confidence_\scale(\outputs) = \min\lr(){\confidence_\scale(\outputs_{\text{start}}),\confidence_\scale(\outputs_{\text{end}})}$

\subsection{Deciding confidence score thresholds}\label{sec:confidence-configuration}

The main goal of confidence score thresholds is to determine the level of confidence that is sufficient to reach a predefined accuracy.
The choice of thresholds also implies the data flow pattern from small to giant models. 
The challenges to determining the confidence score thresholds are two-fold: 1) discovering the balance between accuracy and energy footprint.
Although there can be a handful of choices of thresholds, the causal relation between thresholds, accuracy and energy footprint is not known a priori; 
2) The estimation of the energy footprint of the overall data flow becomes a key factor that determines the net energy reduction.

In order to discover the causal relation, 
we adopt a sampling-based search to exhaustively find the most energy-saving thresholds for any given accuracy.
The algorithm generates a curve (accuracy vs energy cost, as illustrated by \myc{2} the \emph{threshold performance graph} in Figure~\ref{fig:confidence-overview}), in which each point represents a set of thresholds for each model and the corresponding energy cost and accuracy of data flow. 
The chosen thresholds are deployed to routers to determine the flow of requests.
Based on the confidence score, the requests with scores below  a threshold at a model in the dataflow are passed to larger models.

Formally, given a list of models $M$, the algorithm searches for a number of threshold sets with associated accuracy and energy cost, \ie $\{\left((t_i, \forall \model_i \in M), a, e\right)\}$, where $(t_i, \forall \model_i \in M)$ is a threshold set and $t_i$ is the threshold for $\model_i$, $a$ is the validation accuracy and $e$ is the energy cost. 

\begin{algorithm}[t]
    \KwIn{$D_v$, $\labels$, $M = (\model_1, \dots, \model_n)$}
    \KwOut{$T = \{((t_i,\ \forall m_i \in M), a, \energy) \}$, $T_\text{AP}, T_\text{EO}$}
    Initialize search space $[0,1]^{m-1}$\;
    \Repeat{no new $(a, \energy)$}{
      Randomly pick samples $K$ from search space\;
      Compute $a$ on $D_v$ and $\energy=\Sum{i}{}{\rho_i\energy_{i}}, \forall k \in K$\;
      \If{$\exists k \in K$ $a_{\model_{n-1}} \le a \le a_{\model_n}$}{
        Repeat line 3-4 on serach space $[k-\epsilon, k+\epsilon]$\;
      }
    }
    $T_\text{AP} = \Min[\energy](e_{\model_n}, \energy,\{t_i,\ \forall m_i \in M\})$\;
    $T_\text{EO} = \Max[\energy''](a, \energy,\{t_i,\ \forall m_i \in M\})$, where $a_{\model_{n-1}} \le a$\;
    \caption{Confidence Score Threshold Search}
    \label{algo:threshold-searching}
\end{algorithm}

Algorithm~\ref{algo:threshold-searching} 
constructs causation between thresholds, energy cost and accuracy.
In Step 1-7, we take random samples on all possible choices of thresholds and compute the outcome.
Since it is desirable to produce high accuracy that approaches the best model, we sample more at the interval around the current sample point $k$, \ie $[k-\epsilon, k+\epsilon]$, where $\epsilon$ is a small, empirically chosen value,  in step 5-6 to find more accurate thresholds and smooth the curve. 

In addition to providing a threshold performance graph, \sys further provide two default configurations that choose appropriate confidence score thresholds:

\mypar{Accuracy-Preserving (AP)}
In the AP mode, $\text{Acc}_\text{total}$ is equal to $\text{Acc}(\model_n)$, meaning the threshold is chosen such that the overall accuracy by passing requests through the system is the same as the best model available in construction.

\mypar{Energy-Optimization (EO)}
In the EO mode, we choose thresholds with largest energy saving gain, with the least amount of accuracy drop.
In order to find such thresholds, the second derivative of the threshold performance graph is calculated. 
The point with the largest derivative represents the choice of confidence.

\subsection{Skipping unconfident DNNs}
\label{sec:jump_connection}


Although we have a handful of system configurations to construct the data flow, the number of models in the chained data flow can be large. For example, GPT has eight different versions available. If the data flow is constructed with all these models, the latency through the sequence of models can be unacceptably significant and may further worsen the energy footprint. The challenge here is to design a routing strategy along with a confidence score threshold to route unconfident requests directly to the model that can output a confident prediction.

The basic idea is to route less confident requests to larger models. 
For example, the request with the lowest confidence should be routed to the largest model (\ie $p_3$ in Figure~\ref{fig:confidence-overview} \myc{3} with confidence score $c_3=0.2$ skips the medium model).
And the requests with a confidence score close to the threshold are routed to the successor model (\ie $p1$ with confidence score $c_1=0.7$).
The intuition here is that since we have established the correlation between confidence score and its probability of correctness, large models overall produce higher confidence scores than smaller ones.

For each model, the gain is defined as the energy saved by avoiding the use of a larger model minus its own inference energy. Models with positive gain are retained, ensuring overall energy efficiency.
Once routing is determined, we estimate the likelihood of queries being confidently processed by each model. If a model lacks sufficient confidence, it adds latency and energy consumption without providing client responses. Such models are excluded from deployment, with their queries redirected to larger models to maintain accuracy.

\begin{algorithm}[t]
    \KwIn{$M = (\model_1, \dots, \model_n)$, $C, T$}
    \KwOut{updated $M$ with models removed, skip thresholds $T_\text{skip}$}
    Define function $\mathrm{energy\_benefit} (\model_i)=(\rho_{i}-\rho_{i+1})\energy_{i+1} - \rho_{i}\energy_{i}$\;
    \Repeat{$\forall \model_i \in M$ $\mathrm{energy\_benefit}(\model_i)>0$}{
     $\exists \model_i \in M$ $\mathrm{energy\_benefit}(\model_i)<=0$, remove model from $M$\; 
     Run Algorithm~\ref{algo:threshold-searching} again to find new $T$\;
     $\forall t^{skip}_i \in T_\text{skip},\ t^{skip}_i = \text{LogUniform}(0, t_i),\ t_i \in T$\;
    }
    \caption{Skip Connection Configuration}
    \label{algo:threshold-skip}
\end{algorithm}

Algorithm~\ref{algo:threshold-skip} determines the skip thresholds for online routing and data flow cut for offline deployment.
In step 2, we find a model at a time that has no energy gain but only introduces extra latency. 
Note that the model is scanned from small to large to preserve overall accuracy.
In step 3, we recompute all thresholds since naively passing the proportion of data to a larger model may bring extra latency in request routing.
Finally, we introduce a simple but effective construction for skip connection.
The confidence interval below thresholds is uniformly partitioned into the number of successor models.

\section{Hybrid Serving Dataflow Planner}\label{sec:placement}


\subsection{Optimizing dataflow throughput}
\label{sec:feasible_placement}


The goal of dataflow planning aims at maximizing the throughput for hybrid dataflow. 
Resolving this problem is nontrivial because: 1) the planning has to consider memory constraints where giant models are required to be partitioned across multiple GPUs.
With energy efficiency in mind, the model partitioning planing need to work together with multiplexing models on one GPU;
2) requests may pass through different numbers of models causing heterogeneous data transmission overhead between models.

The primary component of the objective function is the latency overhead from data transmission between models and partitions. Under ideal conditions, where performance degradation arises solely from inter-model (small to large) and intra-partition communication, minimizing the data transmission overhead $\transmission$ directly maximizes throughput.
The value of $\transmission$ depends on the communication path, such as network transmission, remote procedure calls, or direct memory access (DMA). Different channels exhibit varying latency; for instance, local DMA offers at least $10\times$ the bandwidth of network transmission. Overall, $\transmission(g, g')$ is abstracted at the GPU level, as GPU communication underpins the data path.
This overhead can be modeled using integer (binary) programming. Let $x_i^g \in \{0,1\}$ indicate whether model $\model_i$ is placed on GPU $g$. The overhead between two models on GPUs $g$ and $g'$ is then given by
$\transmission(g, g') x_i^g x_j^{g'}$.
Aggregating pairwise overhead terms yields the total cost, which Eq.~\eqref{eq:mp-fix} aims to minimize.
\begin{align}
    \Min\ &  \Sum{i,g}{}{\lr(){
        \Sum{\forall g', j \neq i}{}{\transmission(g,g') x_{j}^{g'}x_{i}^{g}}
    }}\label{eq:mp-fix}
\end{align}
This formulation is naturally constrained on the memory capacity of GPUs and also each model instance must appear exactly once in the planning.
Assuming $\memory_{i}$ is the active memory for model $\model_i$, and $\memory^g$ is the total memory of GPU $g$, Eq~\ref{eq:mp-fix-constraints} provides the constraint for the optimization problem.
\begin{align}
    \text{s.t.}\    & \Sum{g}{}{x_i^g}=1,\ \forall g \ \Sum{i}{}{\memory_{i}x_{i}^{g}} \le \memory^g\label{eq:mp-fix-constraints}
\end{align}
To extend this equation to support model partition, we can treat each partition as a ``standalone'' model in the dataflow construction. The request will pass from the first partition to the end as usual. The only difference is that there is no skip connection and router in between. Such a feature can be achieved by muting $\transmission$ (\ie $\transmission=0$) for any successor model other than last partition. 


\subsection{Avoiding serving overloads in GPUs}
\label{sec:avoid_overload}

Overutilizing computational resource is highly likely to cause contention.
Such phenomena is much significant in GPU, resulting in extra delays in processing concurrent requests.
Unlike CPU with multiple parallel cores, a GPU can only be occupied by a single process.
GPU execution of inference requests consists of ``kernel'', which is executed sequentially on GPUs.
Under any given inference period, if the GPU is fully occupied by kernels, issuing more requests on that GPU will lead to contention.

Based on the observation, we propose a novel metrics \textit{kernel utilization} to measure the aggregated load on a GPU.
For a given period of time, the metric measures the proportion of time that the GPU is executing kernels. This adds one more constraints the formulation where the sum of kernel utilization on each GPU must not exceed 1.
\begin{align}
    \text{s.t.}\    & \forall g,\ \Sum{i}{}{\utilization_{i}x_{i}^{g}} \le 1\label{eq:mp-load-constraints}
\end{align}

Once the parallel plan is decided by ILP, we also apply a set of post-ILP communication optimizations, such as zero-copy data movement, whenever
applicable, because the latter reduces the number of replicated
tensors and corresponding wait time, while keeping the communication volume the same.

\subsection{Replicating serving-intensive DNNs}
\label{sec:replicating_dnns}




In order to reduce the impact of queuing along the chain of model ensemble, model replications may follow a pattern such that models are always kept busy on incoming requests and zero queuing delay on any intermediate ensemble stage.
We define queue delay at model $\model_i$ is $\latency_i$ and it needs to handle $\rho_i$ proportion of the total data.
We define queuing delay at model $\model$ is $\latency_\model$ and it needs to handle $\rho_\model$ proportion of the total data.
In order to have zero waiting, each model shall have the same throughput by relying on replication, such that 
\begin{equation*}
    \frac{\rho_1\latency_1}{R_1S_1} = \frac{\rho_2\latency_2}{R_2S_2} = \cdots = \frac{\rho_n\latency_3}{R_nS_n},
\end{equation*}
where $R_\model$ is the number of replications for model $\model$.

Thus, the replications $(R_1, R_2, \dots, R_n)$ should be $(\rho_1\latency_1, \rho_2\latency_2, \dots, \rho_n\latency_n)/R_\model$.
We round the real number to the nearest positive integer.
To note that, model parallelism with partitioning also increases throughput for each model by $S_\model$. As a result, $(R_1S_1, R_2S_2, \dots, R_nS_n)$ should be proportional to $(\rho_1\latency_1, \rho_2\latency_2, \dots, \rho_n\latency_n)$.
This further adds constraints to the decision variable during deployment.

Equation~\eqref{eq:mp-fix} considers a subproblem when all model configuration, \ie maximum batch size, number of replication and number of stages in model parallel.
Meanwhile these variables can be part of the model placement decision to search for the most optimal deployment plan.

The computational time of a kernel depends on the size of the data, usually larger batch size leads to longer time spent on each element of computation. 
Also, the communication latency is proportional to the bytes transmitted.
Such that $\utilization$, $\memory$ and $\transmission$ are all functions of batch size.
Since the confidence and threshold determine the proportion of data $\rho_i$ handled by each model $\model_i$, each model processed $\rho_ib$ batch size for a given batch.
Finally we may replace the corresponding function with $\utilization_i(\rho_ib)$, $\memory_i(\rho_ib)$, $\transmission_i(\rho_ib)$.
Such dependency on batch size can be determined on the pass through validation dataset, with selective batch sizes and linear regression.

The support of model parallelism through layer-wise model partition is straightforward.
Each model partition can be treated as an individual model in the chain of ensemble,
while each model occupies a part of memory $\omega_i/S_i$.
The data transmission time $\transmission$ needs to be reassessed according to the size of hidden states.



\subsection{Efficient planning for large-scale clusters}
\label{sec:planning_algorithm}





\begin{algorithm}[t]
    \KwIn{$\eval$, $M$, $\mathbf{P} = \{\rho_i\latency_i\}$}
    \KwOut{$R,S,\{x^g_i\}$}
    \ForEach{$\model_i \in M$}{
        Profile $\utilization_i$ of $\model_i$ under all $b$\;
        Profile $\transmission_i$ duration of $\model_i$ under all $b$\;
        Establish linear function $b\mapsto \utilization_i, \transmission_i$\;
    }
    \ForAll{$R,S\in\mathbbm{Z}^+$}{
        Validate $\norm{R\mathbf{P}}_1 \le \norm{\mathbbm{\omega}}_1$\;
        ILP solver for Equation~\eqref{eq:mp-fix} with constraints\;
    }
    Record $R,S$ with the maximum objective value\;
    \caption{Placement Search}
    \label{algo:placement-searching}
\end{algorithm}

Algorithm~\ref{algo:placement-searching} combines all previously mentioned strategies and describes how critical data is measured.

First, from line 1 to 4, we profile kernel utilization and data transmission time under all batch sizes $b$ for each model. After profiling, linear maps from batch size $b$ to kernel utilization and data transmission time are built up. The time complexity of line 2 and 3 is $O(b)$ while it is $O(b^3)$ for line 4. Therefore, the overall time complexity from line 1 to 4 is $O(|M|b^3)$.

From line 5 - 7, the algorithm enumerates all possible combinations of the number of replicas and the number of partitions, \ie $(R, S)$ and solves the objective equation using a ILP solver based on profiled data for each combination. Line 6 takes $O(g|M|)$ time to validate if the combination exceeds memory constraints. In line 7, an ILP solver takes $O(g|M|)$ time to compute the solution for the knapsack problem. Therefore, the overall time complexity from line 5 - 7 is $O(RSg|M|)$.

We ensure the placement algorithm can find a reasonable parallelism plan in polynomial time, \ie $O(|M|b^3+RSg|M|)$. Considering that $S\le g$, the number of replicas is limited by the memory, and the number of possible batch sizes is about 10s, even in a large-scale serving cluster (\ie 10s DNNs in the dataflow and 1000s GPUs), the time complexity of the algorithm $<O(10^8)$, which can be efficiently solved.

\section{Evaluation}


In this section, we present the evaluation of \sys. Our evaluation
aims to answer the following questions: 
\begin{itemize}
    \item Can \sys preserve high
model inference accuracy as giant DNNs? (\S\ref{sec:eval-accuracy}) 
\item What is the performance of \sys in terms of request processing latency and throughput? (\S\ref{sec:eval-performance})
\item Can \sys reduce energy cost? (\S\ref{sec:eval-energy})
\item What is the overhead of using \sys? (\S\ref{sec:micro-benchmarks}) 
\end{itemize}

\begin{table}[t]
\small
  \centering
  \caption{DNN models on HuggingFace. $\spadesuit$ indicates model distillation. $\heartsuit$ indicates quantization. Model indicates the unique name of this HuggingFace model.}
    \begin{tabular}{|c|l|l|}
    \hline
    Name & Model (\#Parameters, Accuracy) & Notation\\
    \hline
    \hline
    \multirow{4}[2]{*}{T5}  & t5-small-lm-adapt (60M, 78.2\%) & T5-S \\
      & t5-base-lm-adapt (220M, 84.2\%) & T5-M \\
      & t5-large-lm-adapt (770M, 87.1\%) & T5-L \\
      & t5-xl-lm-adapt (3000M, 90.5\%) & T5-XL \\
    \hline
    \hline
    \multirow{7}[2]{*}{GPT}  & distilgpt2 (86M, 23.6\%)  & GPT-XS $\spadesuit$\\
      & gpt2 (124M, 31.1\%) & GPT-S \\
      & gpt2-medium (345M, 34.4\%) & GPT-M \\
      & gpt2-large (774M, 35.4\%) & GPT-L \\
      & gpt2-xl (1558M, 36.9\%) & GPT-XL \\
      & gpt-j-6B (6700M, 42.3\%) & GPT-XXL \\
      & gpt-j-6B-8bit (6700M, 35.5\%)  & GPT-Q $\heartsuit$\\
    \hline
    \hline
    \multirow{5}[2]{*}{ViT}  & vit-tiny-patch16-224 (12M, 74.8\%) & ViT-XS  $\spadesuit$\\
      & vit-small-patch16-224 (45M,  80.8\%)  & ViT-S $\spadesuit$\\
      & vit-base-patch16-224 (86M, 81.2\%) & ViT-M \\
      & vit-large-patch16-224 (307M, 82.3\%) & ViT-L \\
      & FQ-ViT (307M, 81.3\%)  & ViT-Q $\heartsuit$\\
    \hline
    \end{tabular}%
  \label{tab:model-variants}%
\end{table}%

\para{DNN models.} 
We evaluate \sys with a wide spectrum of giant DNN models (as shown in Table~\ref{tab:model-variants}): (i)~Google T5~\cite{DBLP:journals/jmlr/RaffelSRLNMZLL20} is the foundation models for text-to-text generation. There are 4 variants of T5 with parameters from 60 million to 3000 million. These models exhibit accuracy from 78.2\% (T5-S) to 90.5\% (T5-XL) in the GLUE~\cite{DBLP:conf/iclr/WangSMHLB19} multi-label classification task.
(ii)~GPT~\cite{gpt3} is the foundation model for generative inference. There are 7 variants of GPT with parameters from 86 million to 6700 million. These models exhibit accuracy from 23.6\% (GPT-XS) to 35.5\% (GPT-XXL) in the LAMBADA~\cite{DBLP:conf/acl/PapernoKLPBPBBF16} zero-shot next word prediction task. 
(iii)~ViT~\cite{vit} is the foundation model for image and video analytics. There are 5 model variants with parameters from 12 million to 307 million. These models exhibit accuracy from 74.8\% (ViT-XS) to 81.3\% (ViT-L) in the ImageNet~\cite{DBLP:conf/cvpr/DengDSLL009} multilabel classification task.

\para{\sys and baselines.} 
\sys uses PyTorch (v1.11.0), Ray~(v1.9.2) and Triton (r22.21). We report the results of \sys with two configurations: (i)~\sys (AP) chooses the confidence score threshold that preserves the accuracy of the giant DNN, implying its Accuracy-Preserving (AP) purpose. (ii)~\sys (EO) chooses the threshold that optimizes energy efficiency, implying its Energy-Optimization (EO) purpose.

There are two categories of baselines: (i)~Accuracy baselines include model distillation, quantization and the original giant DNN, allowing us to evaluate the accuracy performance of \sys against SOTA
model compression techniques. (ii)~Performance baseline: DeepSpeed~\cite{DBLP:conf/kdd/RasleyRRH20} (version: 0.5.8)
which is the SOTA high-performance system for serving giant DNNs. 
We omit comparison against Nvidia~Triton
and Ray~Serve because \sys is implemented on top of these two technologies and has achieved superior performance. 

\para{Testbed setup.} 
We run experiments on two test-beds: (i)~A 8-GPU server (Nvidia A5000) that has 112 CPU threads
and 1 TB memory. (ii)~A cloud 12-server cluster where each server has a Nvidia T4 GPU, 8 CPU threads and 128 GB memory.

\subsection{Accuracy}\label{sec:eval-accuracy}

\begin{table}[t]
\small
  \centering
  \caption{Request distribution in \sys}
    \begin{tabular}{|l|c|c|}
    \hline
          & \multicolumn{2}{c|}{Request Distribution (Threshold)} \\
    \hline
    Model & \sys (AP) & \sys (EO) \\
    \hline
    \hline
    T5-S  & 0\% (0.98)  & 40\% (0.86) \\
    T5-M  & 24.8\% (0.86)  & 42.3\% (0.71)\\
    T5-L  & 28.2\% (0.78) & 9.2\% (0.72)\\
    T5-XL & 47\% (0.0) & 8.5\% (0.0)\\
    \hline 
     Overall   & 90.5\% & 89.8\%\\
    \hline
    \hline
    GPT-XS & 0\% (1.0)   & 0\% (1.0)\\
    GPT-S & 0\% (0.91)   & 19.5\% (0.40) \\
    GPT-M & 19.8\% (0.51) & 50.9\% (0.13) \\
    GPT-L & 20.3\% (0.36) & 12.9\% (0.14) \\
    GPT-XL & 0\% (0.78)  & 0.1\% (0.42) \\
    GPT-XXL & 59.9\% (0.0) & 16.6\% (0.0)\\
    \hline 
     Overall   & 42.3\% & 37.2\%\\
    \hline
    \hline
    ViT-XS & 59.6\% (0.71)  & 59.6\% (0.71)\\
    ViT-S & 5.1\% (0.89)  & 31.0\% (0.34)\\
    ViT-M & 18.2\% (0.58) & 4.5\% (0.33)\\
    ViT-L & 17.1\% (0.0) & 4.9\% (0.0)\\
    \hline 
     Overall   & 82.3\% & 81.7\%\\
    \hline
    \end{tabular}%
  \label{tab:distribution}%
\end{table}%

\begin{table*}[t]
  \centering
  \caption{Accuracy of \sys and baseline techniques.}
    \begin{tabular}{|c|c|c|c|c|c|c|c|c|c|c|}
\cline{1-3}\cline{5-7}\cline{9-11}    \multicolumn{2}{|c|}{T5} & Accuracy &          & \multicolumn{2}{c|}{GPT} & Accuracy &          & \multicolumn{2}{c|}{ViT} & Accuracy \\
\cline{1-3}\cline{5-7}\cline{9-11}    Giant Model & T5-XL    & 90.5\%   &          & Giant Model & GPT-XXL  & 42.3\%   &          & Giant Model & ViT-L    & 82.5\% \\
\cline{1-3}\cline{5-7}\cline{9-11}    Distillation & N/A      & N/A      &          & Distillation & GPT-XS   & 23.6\%   &          & Distillation & ViT-XS   & 74.5\% \\
\cline{1-3}\cline{5-7}\cline{9-11}    Quantization & N/A      & N/A      &          & Quantization & GPT-Q    & 35.5\%   &          & Quantization & FQ-ViT   & 81.3\% \\
\cline{1-3}\cline{5-7}\cline{9-11}    \multicolumn{2}{|c|}{\sys (AP)} & 90.5\%   &          & \multicolumn{2}{c|}{\sys (AP)} & 42.3\%   &          & \multicolumn{2}{c|}{\sys (AP)} & 82.5\% \\
\cline{1-3}\cline{5-7}\cline{9-11}    \multicolumn{2}{|c|}{\sys (EO)} & 89.8\%   &          & \multicolumn{2}{c|}{\sys (EO)} & 37.2\%   &          & \multicolumn{2}{c|}{\sys (EO)} & 81.7\% \\
\cline{1-3}\cline{5-7}\cline{9-11}    \end{tabular}%
  \label{tab:accuracy}%
  \vspace{-0.2in}
\end{table*}%


We evaluate \sys'’s accuracy under two configurations -- Accuracy-Preserving (AP) and Energy-Optimized (EO) -- on the 8-GPU server (Table~\ref{tab:accuracy}).

\mypar{Effects of accuracy-preservation} In all the DNN models (\ie T5, GPT and ViT), \sys (AP) can achieve
the same accuracy as the giant DNNs (the ideal case) while both model distillation and quantization largely reduce the inference accuracy. Specifically, model distillation reduces the accuracy from 37.2\% to 23.6\% in GPT and from 81.7\% to 74.5\% in ViT. Quantization reduces the accuracy from 37.2\% to 35.5\% in GPT and from 81.7\% to 81.3\% in ViT. Note that both model distillation and quantization have not been applied to T5 yet on HuggingFace. This is because these techniques are model-specific, making them difficult to be used as a generic energy-saving technique. In contrast, \sys treat the DNNs as blackboxes and can be applied in general, making them easy to be adopted.

\sys's high inference accuracy performance does not compromise its energy efficiency. As shown in Table~\ref{tab:distribution} which shows the distribution of requests processed by different DNNs, in serving T5, \sys (AP) delegates 24.8\% requests to T5-M (which is 95\% smaller than the giant model: T5-XL) and 28.2\% requests to T5-L (which is 74\% smaller than T5-XL). \sys thus reduces the 53\% requests that will reach T5-XL, thus reducing the number of its replicas and the associated energy cost. We observe a similar improvement in energy cost in GPT and ViT. For example, as shown in Table~\ref{tab:distribution}, in serving ViT, 59.6\% requests are delegated to ViT-XS, 5.1\% are delegated to ViT-S and 18.2\% are delegated to ViT-M. As a result, 82.9\% requests will not reach the energy-intensive ViT-L (which is 24 $\times$ larger than ViT-XS). This indicates that \sys can greatly reduce the workload on giant DNNs by delegating the requests to significantly smaller DNNs.

\mypar{Effects of energy-optimization}
We also evaluate the accuracy cost caused by choosing the confidence threshold for saving energy.
As shown in Table~\ref{tab:accuracy}, in serving T5 and ViT, \sys (EO) achieves accuracy performances that are very close to corresponding giant DNNs (\ie 89.8\% vs. 90.5\% in T5 and 81.7\% vs. 82.5\% in ViT). These accuracy costs are significantly smaller than those incurred in model distillation and quantization. By introducing a small drop in accuracy, \sys delegates a significant proportion of requests to small DNNs. As shown in Table~\ref{tab:distribution}, in the case of T5,
\sys (EO) delegates 91.5\% requests to smaller DNNs, 38.5\% more than the \sys (AP).
In the case of ViT, \sys (EO) delegates 95.1\% requests to smaller DNNs, 12.2\% more than the \sys (AP).

We make an interesting observation in the results of GPT. \sys (AP) has an accuracy cost of 5.1\%. Though smaller than the costs incurred by distillation (18.7\%) and quantization (6.8\%), this accuracy cost is more significant compared to those incurred in T5 and ViT. A key reason for this is: the second largest DNN (GPT-XL) is poorly trained. As shown in Table~\ref{tab:distribution}, GPT-XL covers less than 0.1\% requests. This indicates the need for re-training this second largest DNN for the new given dataset, making it more capable of processing requests and thus can act as a capable cover for the giant DNN (GPT-XXL).

\subsection{Performance}\label{sec:eval-performance}

We then evaluate the performance of \sys in the 12-server cluster. The evaluation is grouped into
those for measuring (i) the averaged request processing latency,
(ii) the tail (99.9\%) latency, and (iii) the request throughput.

\begin{figure}[t]
    \centering
    \begin{subfigure}{.8\linewidth}
        \includegraphics[width=\linewidth]{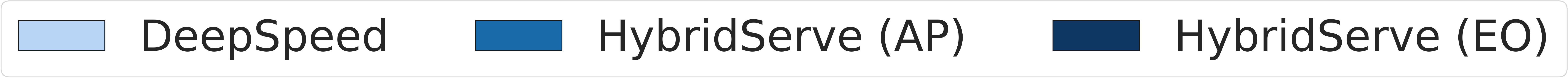}
    \end{subfigure}
    \begin{subfigure}{.48\linewidth}
       \includegraphics[width=.8\linewidth]{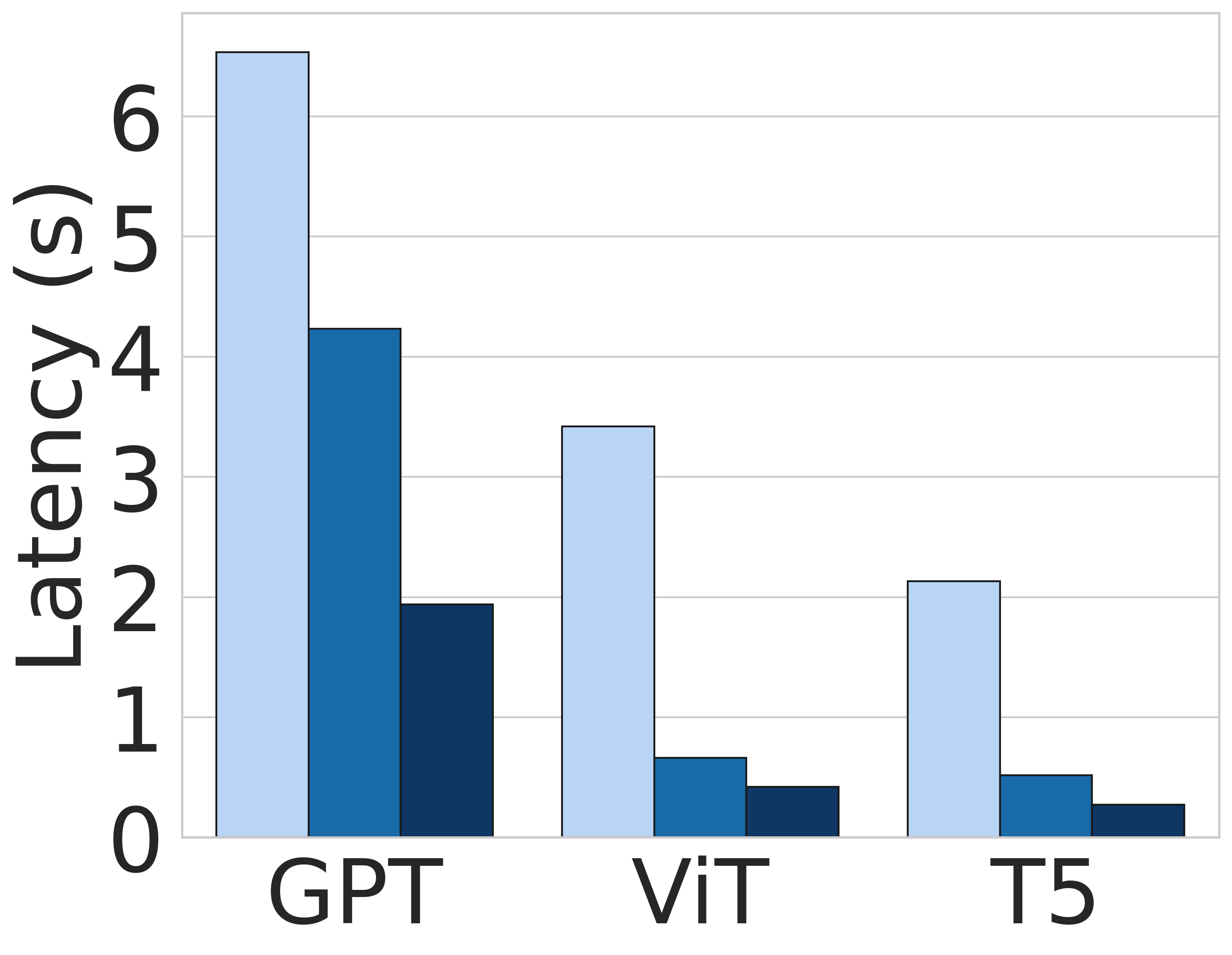}
       \caption{Average latency.}
    \end{subfigure}
    \begin{subfigure}{.48\linewidth}
       \includegraphics[width=.8\linewidth]{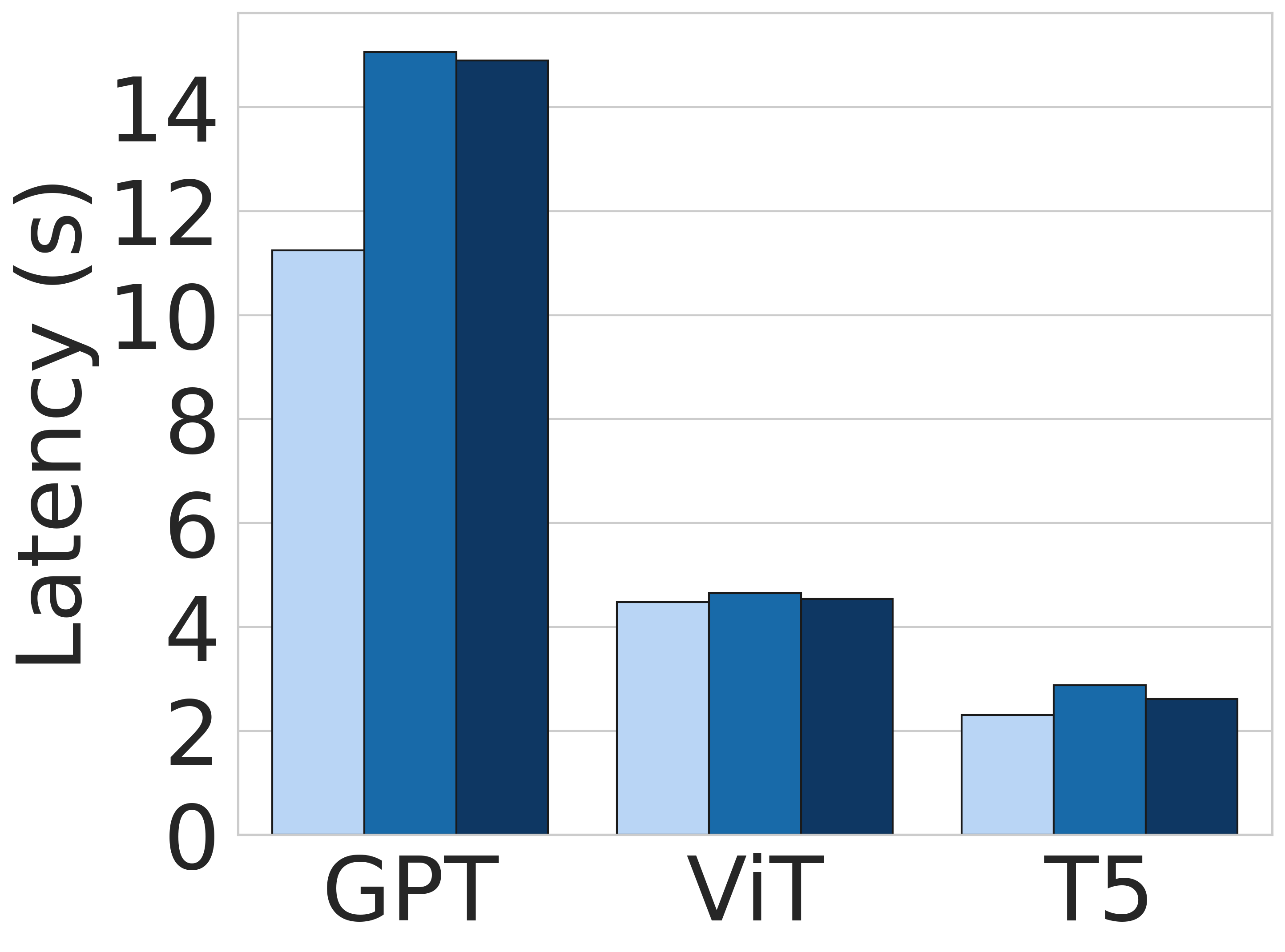}
       \caption{Tail latency.}
    \end{subfigure}
    \caption{Latency performance}
    \label{fig:eval-latency}
\end{figure}

\mypar{Average latency}
Figure~\ref{fig:eval-latency} (a) shows the averaged latency in processing DNN inference requests.
For ViT, since most requests are delegated to smaller DNNs, \sys can achieve an averaged latency of 610 ms (if using the accuracy-preserving configuration) and 385 ms (if using the energy-optimization configuration). These latency results are an order of magnitude better than DeepSpeed which spends 3420 ms in processing a request in average. This latency improvement comes from the fact that: (i) \sys largely benefit from a cautious usage of computational-efficient small DNNs and (ii) \sys reduces the need for transmitting data over the cloud network. We observe similar improvement in averaged latency in the GPT and T5 models, indicating the effectiveness of \sys in low-latency services. 

\mypar{Tail latency} The tail latency is a major performance cost incurred by \sys. Since a request first needs to pass through a small DNN and then reach larger DNNs, \sys costs at least one extra hop compared to DeepSpeed which directly uses the giant DNN for processing this request. 
As shown in Figure~\ref{fig:eval-latency} (b), the tail latency of \sys is within 1.2\% in ViT and 2\% in T5. This is because the skip connections in \sys are effective in reducing the number of extra routing hops and almost all requests can directly jump to the giant DNNs. Since the smallest used DNNs are 98\% and 96\% are smaller than their largest ones in T5 and ViT, respectively. The latency overhead of passing through these small DNNs thus becomes marginal. 
The latency overhead however is more significant in GPT: \sys's tail latency is 18\% longer than DeepSpeed. This is because the smallest used DNN (GPT-M) is still large in size (8\% of the original size). We anticipate this tail latency overhead can be further reduced by adopting more powerful small DNNs (such as a re-trained GPT-XS). 

\begin{figure}[t]
    \centering
    \includegraphics[width=.85\linewidth]{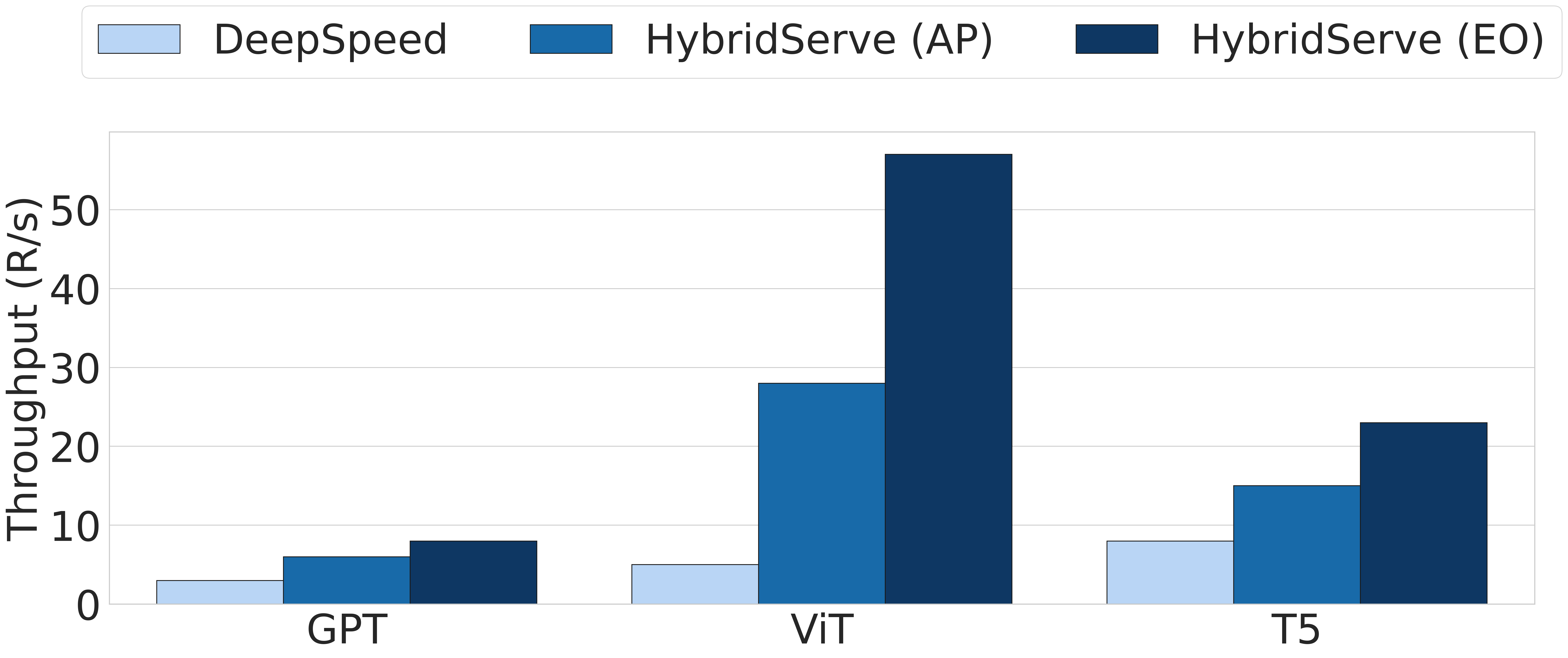}
    \caption{Throughput performance}
    \label{fig:eval-throughput}
\end{figure}

\mypar{Request throughput} Figure~\ref{fig:eval-throughput} shows the throughput of processing requests in \sys and DeepSpeed.  \sys is 
8.9 $\times$, 2.9 $\times$ and $1.7\times$ faster than DeepSpeed in serving ViT, T5 and GPT, respectively, reflecting \sys as an resource-efficient choice for serving giant DNNs.

\subsection{Energy cost}\label{sec:eval-energy}
Evaluating the energy cost of \sys poses a unique challenge: there is no widely accepted metric for evaluating the energy cost of a distributed DNN serving system. We thus design a metric -- Joules per request. To compute this metric, we first measure the energy consumed by a GPU in a time frame. Specifically, the energy is measured every 100ms using the Nvidia NVML library. We accumulate the periodic measurements and divide the accumulated energy by the number of requests that have arrived in this time frame.
The energy cost also includes idle GPU energy consumption.


\begin{figure*}[t]
    \centering
    \begin{subfigure}{.3\linewidth}
        \includegraphics[width=.85\linewidth]{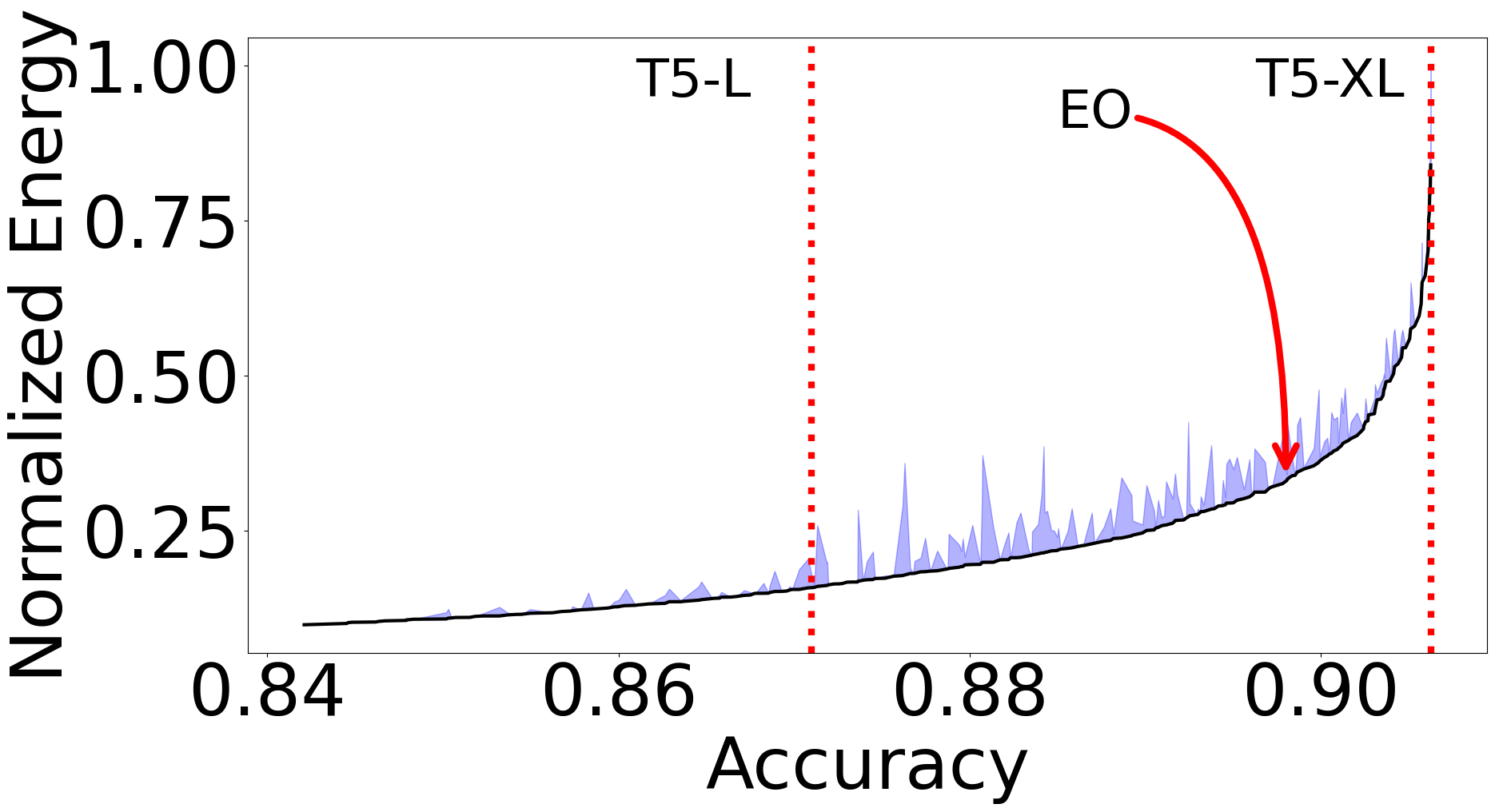}
    \end{subfigure}
    \begin{subfigure}{.3\linewidth}
        \includegraphics[width=.85\linewidth]{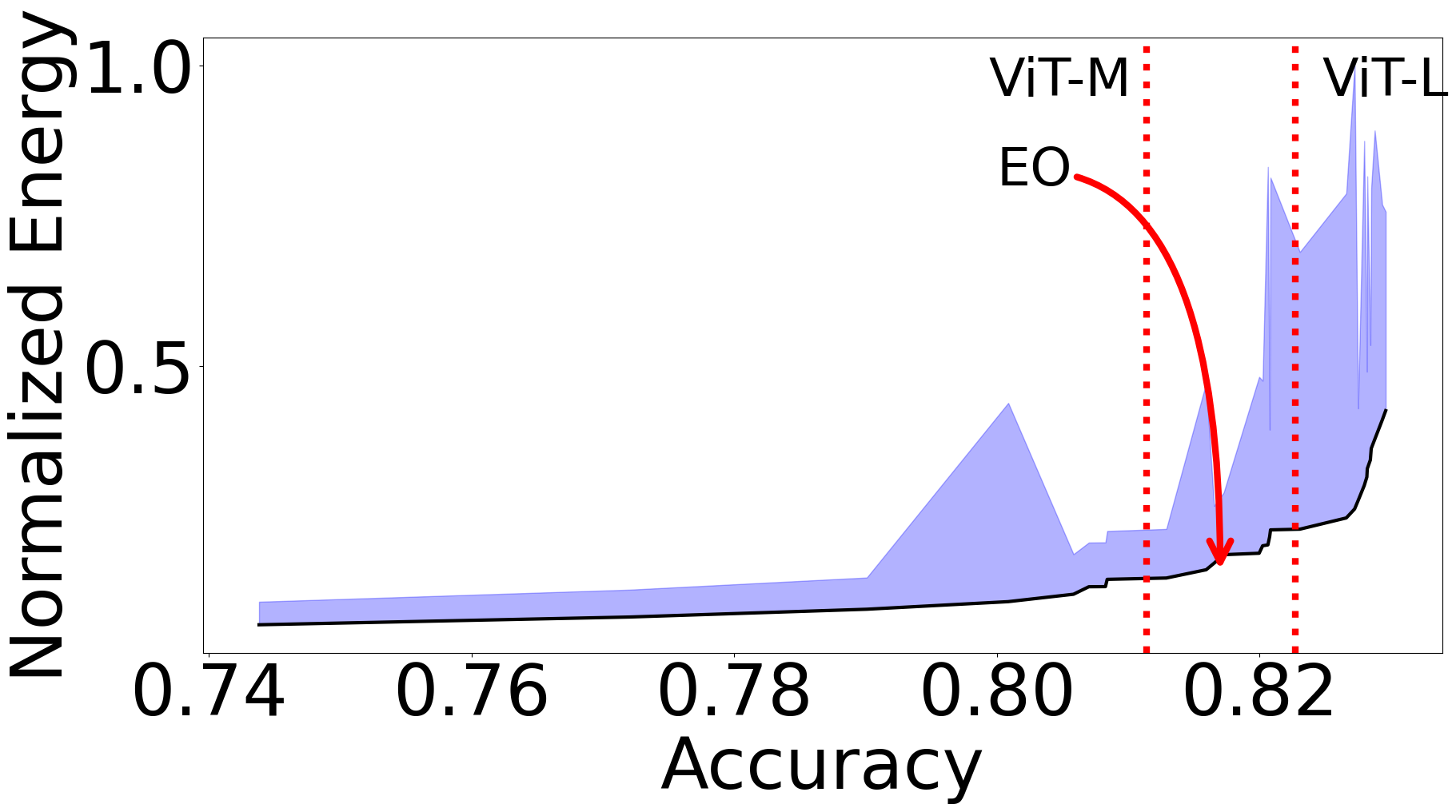}
    \end{subfigure}
    \begin{subfigure}{.3\linewidth}
        \includegraphics[width=.85\linewidth]{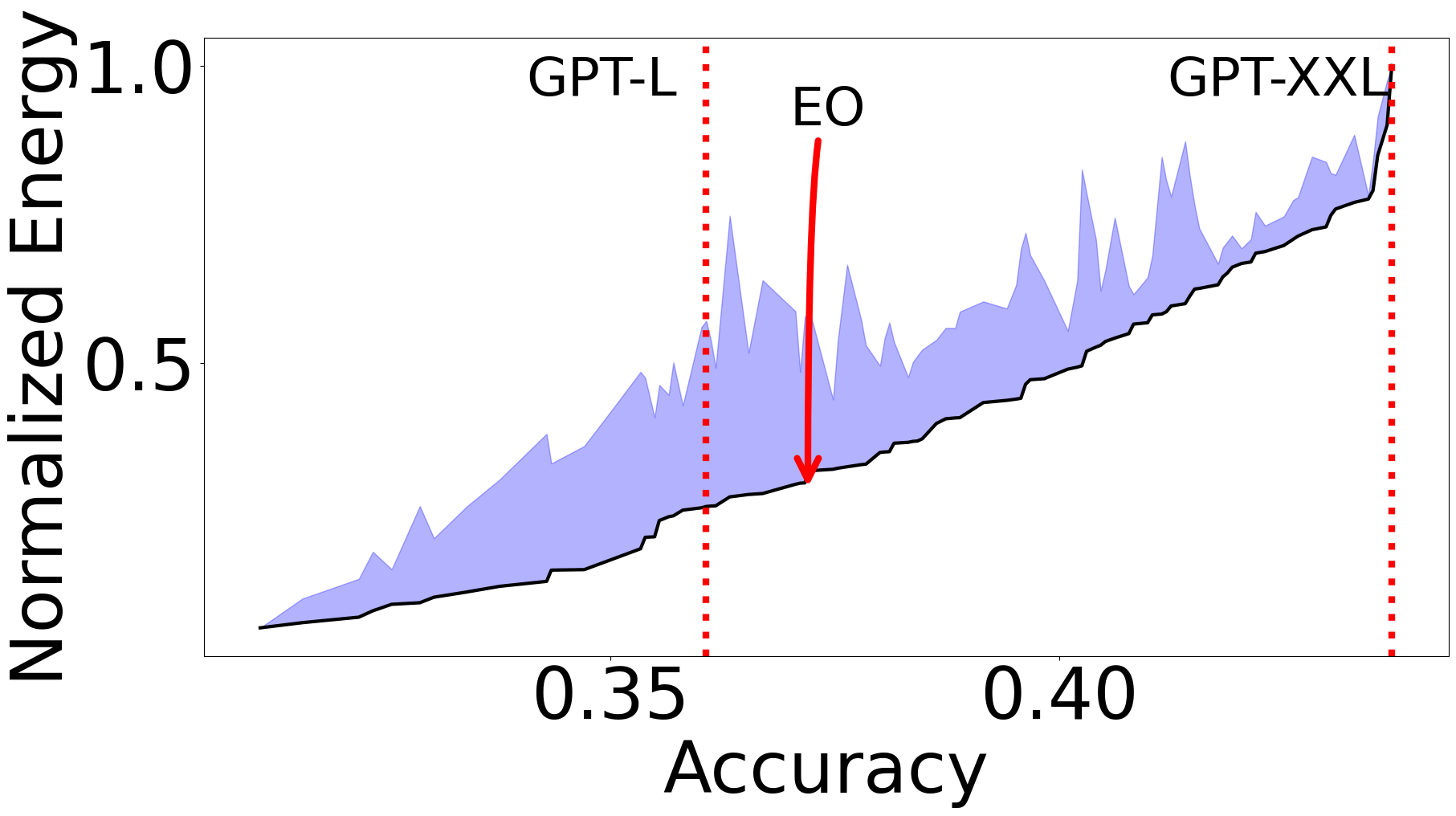}
    \end{subfigure}
    \caption{Relation between normalized energy cost and inference accuracy. Each accuracy point is associated with a configuration for confidence score thresholds. Shaded areas denote the energy costs achieved by different threshold configurations leading to the same accuracy.  The vertical lines indicate the accuracy achieved by different DNN models. T5, ViT, GPT from left.}
    \label{fig:energy_vs_accuracy}
    \vspace{-0.2in}
\end{figure*}

\mypar{Energy saving} Table~\ref{tab:eval-energy} shows the energy cost of \sys and DeepSpeed. \sys (AP) achieves 14$\times$, 9$\times$
and 1.9 $\times$ lower energy cost compared to DeepSpeed in serving ViT, T5 and GPT, respectively. An interesting observation is that: the improvement in energy cost is even higher than the one in averaged latency (9.1 $\times$). This is because \sys can pack multiple DNNs into a single GPU, thus yielding better resource efficiency which is not reflected in request processing latency. This indicates the need for enabling DNN multiplexing and avoiding GPU overloads in \sys's dataflow planner. 

\begin{table}[t]
    \centering
    \caption{Energy savings comparison (per request in mWh).}
    \begin{tabular}{|c|c|c|c|}
    \hline
        System      & GPT   & ViT   & T5    \\
    \hline
        DeepSpeed   & 130   & 115    &  80   \\
        \sys (AP)   & 72    & 8     &  9    \\
        \sys (EO)   & 31    & 6     &  7    \\
    \hline
    \end{tabular}
    \label{tab:eval-energy}
\end{table}

Using \sys (EO) can achieve even more energy savings. As we can see from Table~\ref{tab:eval-energy}, compared to DeepSpeed, \sys (EO) can achieve 19.8 $\times$ and 11.4 $\times$ energy saving in serving ViT and T5 while only incurring 0.4 \% and 0.6 \% accuracy costs. 

\mypar{Energy saving vs. Accuracy cost} We further study the relation between energy-saving and the accuracy cost in \sys. 
Figure~\ref{fig:energy_vs_accuracy} show such relations for T5, ViT and GPT. As we can see, the relations are often superlinear: it costs more energy to improve accuracy if the accuracy is already high. This indicates that exists a sweet point that can significantly save energy by only compromising a little accuracy. Such a sweet point can be found by \sys (EO). As we can see, the confidence threshold is chosen by efficiently searching potential threshold configurations. 
\sys can return the optimized threshold in polynomial time.

\sys (EO) has a bounded accuracy cost. As shown by Figure~\ref{fig:energy_vs_accuracy}, its resulting accuracy is always better than the second-largest DNNs. This accuracy cost bound ensures \sys can yield better accuracy performance than model distillation and quantization techniques.

\subsection{Overheads}\label{sec:micro-benchmarks}

Finally, we evaluate the overheads of adopting \sys. 
There are two major overheads: (i)~The time of 
constructing and planning the hybrid model serving dataflow
and (ii)~The time incurred by confidence based request routing:


\begin{figure}[t]
    \centering
    \begin{minipage}{0.48\linewidth}
        \includegraphics[width=\linewidth]{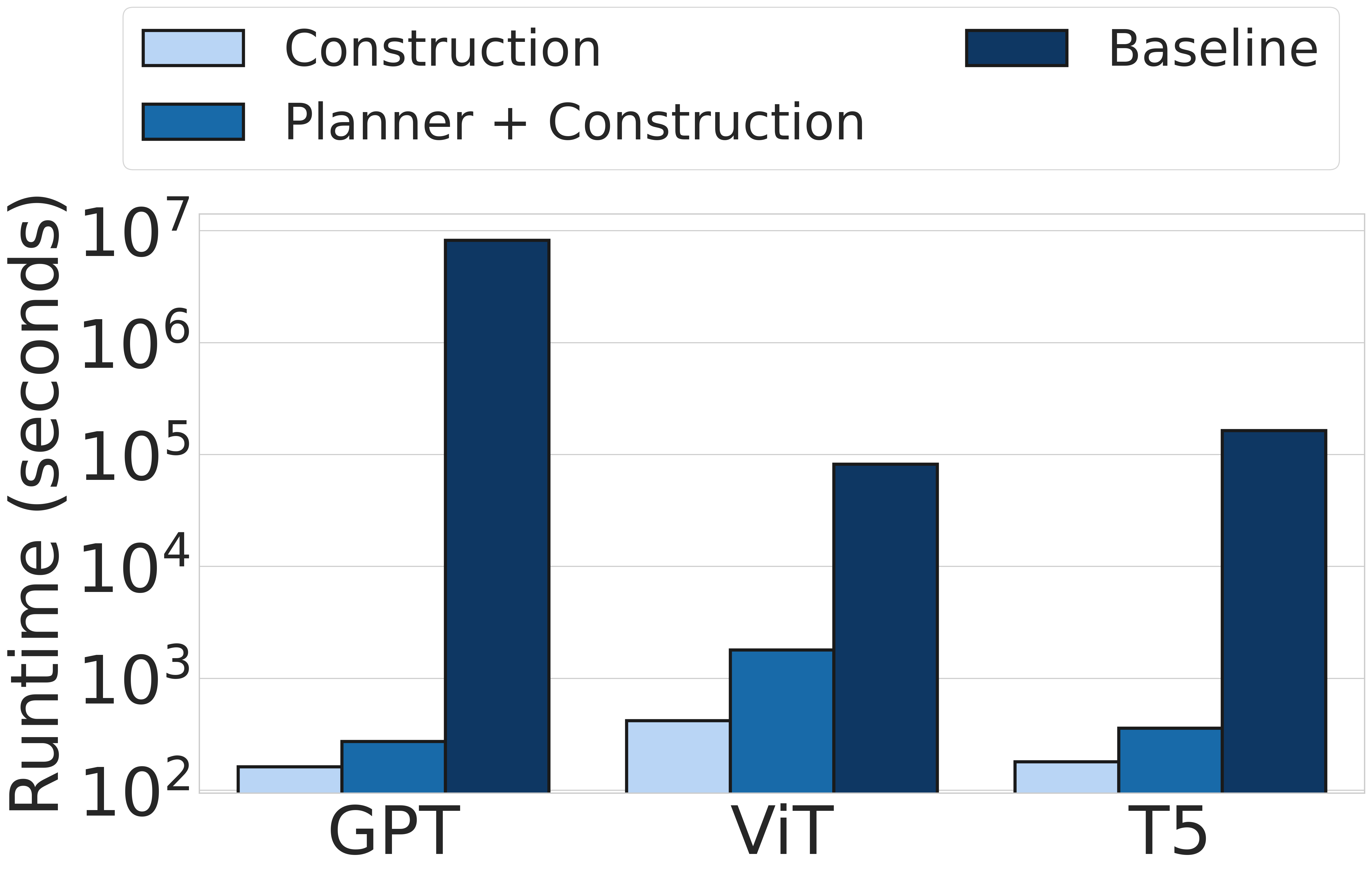}
        \caption{Dataflow construction and planning.}
        \label{fig:dataflow_overhead}
    \end{minipage}
    \begin{minipage}{0.48\linewidth}
        \includegraphics[width=\linewidth]{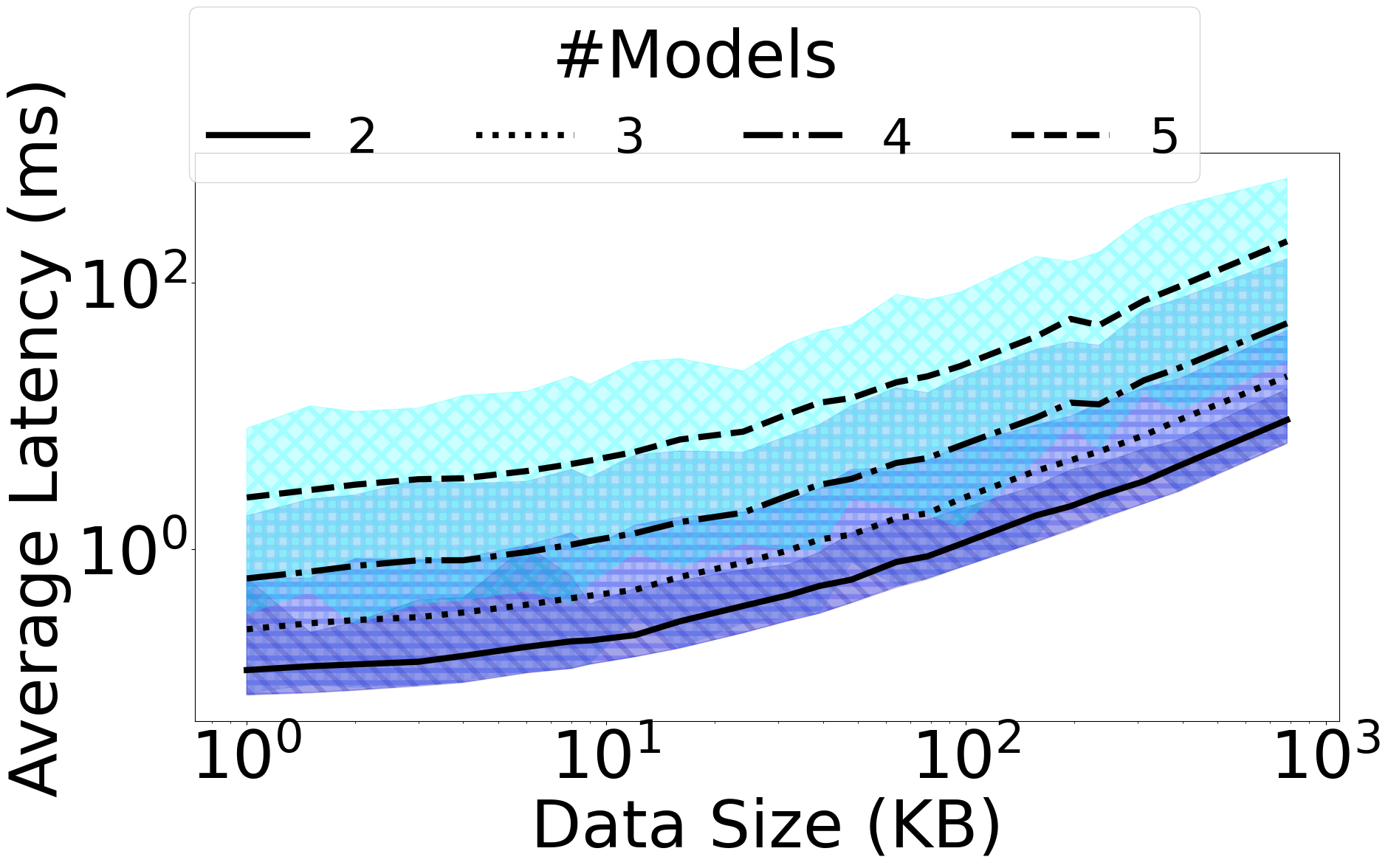}
        \caption{Request routing}
        \label{fig:routing_overhead}
    \end{minipage}
\end{figure}

\mypar{Dataflow construction and planning}
Figure~\ref{fig:dataflow_overhead} reports the time of constructing and planning the dataflow in \sys. 
Compared to training a giant DNN for one epoch (denoted by Baseline), the construction time of dataflow is 3\%, 6\% and 15\% shorter than the time spent in training DNN models: GPT, T5 and ViT, respectively. We further add the time in running the dataflow planner. 
In Figure~\ref{fig:dataflow_overhead}, the total time of dataflow construction and planning is still significantly shorter than training the models for one epoch. \sys can finish the dataflow construction and planning in 45 seconds, 78 seconds and 437 seconds in the case of GPT, T5 and ViT, respectively.
This indicates the low overhead of adopting \sys even in a large DNN serving cluster.

\mypar{Request routing}
Figure~\ref{fig:dataflow_overhead} shows the averaged latency of processing the predictions in varied sizes in a \sys request router. The Data Size denotes the size of the prediction vector in bytes (from 1K byte to 1M bytes). We also vary the number of DNNs included in the hybrid serving dataflow. The number of DNNs is proportional to the complexity of passing through the routing table in the router. As we can see, even when processing large prediction vectors that have 1M bytes, \sys can complete the routing in around 100 ms (the predictions are assigned to 5 DNN models). In practice, prediction vectors are much smaller than 1 MB. For example, the size of GPT's prediction vector is 50K bytes which can be routed within 1 ms.

\section{Related Work}


\mypar{Model parallelism}
\review{
Method to partition the model~\cite{alpaserve} is typically determined by users according to performance requirement~\cite{serverlessllm,moe-infinity}.
\sys introduces a hybrid model serving dataflow which works on top of the model parallelism with additional consideration for sharing the GPU compute and memory.
The scale of dataflow dependency and routing is mainly between models, rather than being fine-grained on GPU operations for a single model.
}


\mypar{GPU sharing}
Multiplexing GPUs among multiple model instance can be implemented using temporal sharing~\cite{reef}, spacial sharing~\cite{clipper} or both~\cite{orion}.
\sys provides a performance guarantee by optimizing the placement of the hybrid serving dataflow, while the runtime sharing is a complementary optimization for better GPU utilization.




\mypar{DNN prediction confidence}
Theoretical methods~\cite{DBLP:conf/nips/Lakshminarayanan17, DBLP:conf/nips/SnoekOFLNSDRN19} have been developed to strengthen the confidence score on a non-iid dataset.
Recently, confidence score has also been proved effective on model ensembles~\cite{wang2022wisdom,DBLP:conf/iclr/HendrycksG17}.
\sys, on the other hand, utilizes the features of confidence outputs to guide the routing of dataflow across all models to achieve performance improvement and energy saving.



\mypar{Cloud-edge model inference collaboration}
Model collaboration between device and cloud has been introduced to speed up inference performance of latency and accuracy.
Such strategy often involves model partitioning across device-cloud~\cite{DBLP:conf/asplos/KangHGRMMT17,DBLP:conf/icdcs/FangJZ19,DBLP:conf/icdcs/Teerapittayanon17} or training~\cite{DBLP:conf/kdd/YaoWJHZY21}. 
Even without modification to model parameters, existing collaborative systems often treat the model as a white box~\cite{DBLP:conf/icdcs/JeongLSYM20} .
\sys is able to work with any blacked-boxed model without customization of the model structure. 
We are fundamentally different from any model offloading and partitioning methods since \sys enables a hybrid usage of standalone models.


\section{Conclusions}

We have introduced \sys, an energy-efficient serving system with giant DNNs.
\sys explores a novel design space for DNN serving systems that harness varied sized DNNs together for a cloud-based AI service to reduce the energy consumption of serving with giant DNNs. \sys design brings together a hybrid model serving dataflow and a hybrid serving dataflow planner. Experimental evaluation using a prototype implementation of \sys shows that it can significantly outperform (up to 19.8 $\times$) state-of-the-art DNN model serving systems, including DeepSpeed, in terms of energy efficiency while offering high inference accuracy similar to serving solely with giant DNNs.

\bibliographystyle{ieeetr}
\bibliography{example}

\end{document}